# The Need and Status of Sea Turtle Conservation and Survey of Associated Computer Vision Advances

Aditya Jyoti Paul [1,2]

1. Cognitive Applications Research Lab,
India.

2. Department of Computer Science and Engineering,
SRM Institute of Science and Technology,
Kattankulathur, Tamil Nadu – 603203, India.
aditya_jyoti@carlresearch.org

*Abstract*— For over hundreds of millions of years, sea turtles and their ancestors have swum in the vast expanses of the ocean. They have undergone a number of evolutionary changes, leading to speciation and sub-speciation. However, in the past few decades, some of the most notable forces driving the genetic variance and population decline have been global warming and anthropogenic impact ranging from large-scale poaching, collecting turtle eggs for food, besides dumping trash including plastic waste into the ocean. This leads to severe detrimental effects in the sea turtle population, driving them to extinction. This research focusses on the forces causing the decline in sea turtle population, the necessity for the global conservation efforts along with its successes and failures, followed by an in-depth analysis of the modern advances in detection and recognition of sea turtles, involving Machine Learning and Computer Vision systems, aiding the conservation efforts.

## I. Introduction

Sea turtles are some of the most majestic creatures of the ocean. They have become an emblem of marine wildlife conservation across the globe. Not only are they one of the most widely recognized and iconic sea creatures but also are one of the oldest, the first creatures of the modern sea turtle species have been around for over a 100 million years.

Rhodin et al. [1] , in the latest taxonomic revision in 2017 defined tortoises as a terrestrial sub-group within turtles (Chelonians). This implies that all tortoises are turtles but not vice versa [2]. In fact, 'turtles' is an umbrella categorization for about 360 currently recognized extant and extinct species of the order Testudines which includes turtles, tortoises and terrapins. Non-tortoise turtles are predominantly aquatic creatures including terrestrial, freshwater as well as sea turtles [3]. Decrease in population of turtles at the current rate are much more pronounced than those observed at the Cretaceous-Paleogene border and suggest an upcoming sixth mass extinction within the next few centuries, in this Anthropocene [4].

The remaining article primarily focuses on sea turtles, the existence and population density of which play an important role in the marine biodiversity, and are also considered a barometer for environmental health. This ancient creature has its unique set of characteristics and vulnerabilities, an analysis of its population offers a lens into a lot of important environmental factors like ocean health and biodiversity, along with the anthropogenic impact on our planet [5].

Currently there are seven species of sea-turtles, which can be grouped into six genera. The common names for the species are Loggerhead, Green Turtle, Leatherback, Hawksbill, Kemp's Ridley and Olive Ridley, and this article will analyze their conservation status and Machine Learning (ML) and Computer Vision (CV) advances aiding the same.

The rest of the article is structured as follows. Section II discusses some of the characteristics of the turtle species under study, Sec. III presents their current conservation statuses, Sec. IV examines the threats and impact of various natural and anthropogenic forces, Sec. V analyzes the advances in ML and CV aiding conservation efforts. Sec. VI discusses some ongoing research, followed by some future avenues in Section VII. Sec. VIII serves as the conclusion for this research.

## II. Species of Sea Turtles

### A. Loggerhead (Caretta caretta)

Loggerhead sea turtles possess comparatively larger heads and that gives them their name as shown in Fig. 1. Their powerful jaws enable them to feed on prey with hard shells like conchs and whelks [6,7]. The top shell of an adult is slightly heart-shaped and reddish-brown in color, with at least five pleural scutes and the plastron is yellow [8]. They are the most common sea-turtles in the US coast. Analyzing the data from 437 loggerheads suggested they weigh between 5.2 to 184.6 kg having a mean ± SD of  46.5 ± 21.6 kg, and a straight carapace length (SCL) of upto 104.5 cm [9].

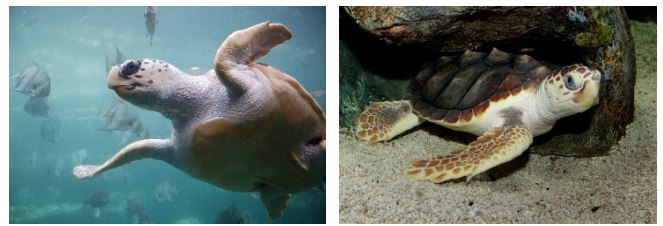

Fig. 1. Loggerhead turtles.

### B. Green Turtle (Chelonia mydas)

Green turtles are endangered but yet found in massive populations in the eastern coasts of Florida. Their carapace is usually heart-shaped and of varying color. They have a small head and single-clawed flippers. They grow to about four feet [10] and adults weigh about 200 kilograms [11]. Some population studies in the northern Great Barrier Reef [12] and also in the northern Persian Gulf and Oman Sea [13], Iran have been discussed. Fig. 2 shows some green turtles in the wild.

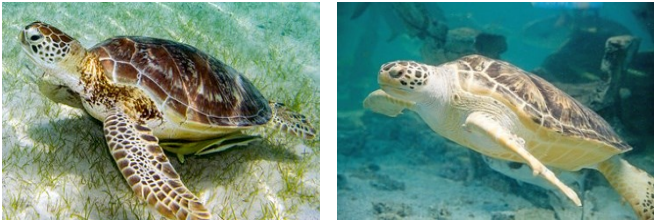

Fig. 2. Green Turtles.

## C. Leatherback (Dermochelys coriacea)

The leatherback is the largest of all sea turtle species. It also dives to the deepest depths of the ocean (over 1000 meters) and is the most migratory species of all sea turtles. The carapace is composed of an agglomeration of osteoderms (small bony plates), covered by firm, elastic and rubbery skin with seven longitudinal keels or ridges [14], instead of the usual keratinous shell [15], as shown in Fig. 3. The skin is mostly black with some light spotting. The adults notably feature a pink spot on the dorsal surface of the head. A toothlike cusp is located on both sides of the gray upper-jaw with an anteriorly hooked lower jaw [16]. Numerous studies [17-20] in the past two decades however indicate that the population is rapidly declining, primarily due to fisheries, harvest of eggs and coastal development.

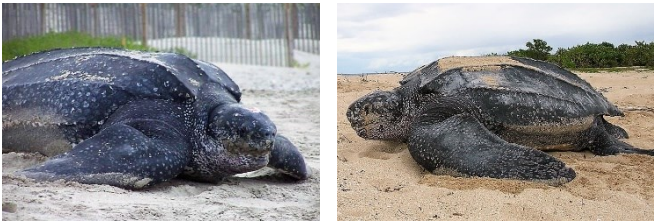

Fig. 3. Leatherback Turtles.

## D. Hawksbill (Eretmochelys imbricata)

Hawksbill turtles are so named for their pointed and narrow beak as shown in Fig. 4. They primarily consume sponges but were also reported to feed on jellyfish, other invertebrates and marine flora [21,22]. Adult hawksbills may reach upto three feet in carapace length and about 140 kilograms in weight, though they typically measure upto 2.5 feet and weigh less than 80 kilograms [23]. They are seldom found in water deeper than 65 feet. Their carapace features a pattern of overlapping scales, giving a distinct serrated look [24]. These patterned and colorful shells, commonly known as 'tortoiseshell', make them very attractive for poachers [25-27], which along with other factors like by-catch and loss of their coral reef habitat, justified its critically endangered IUCN status in 1996 [28] (further explained in Section III). They play an important role in the marine ecosystem, supporting a variety of epibionts [29] and have now spread to sea pastures as a peripheral habitat [30].

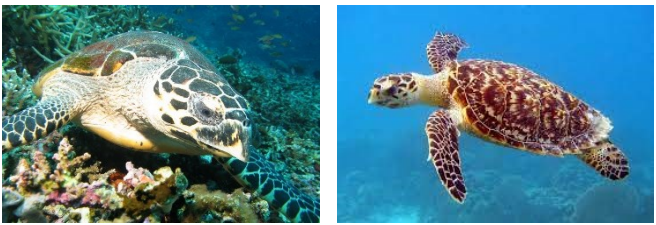

Fig. 4. Hawksbill Turtles.

## E. Kemp's ridley (Lepidochelys kempii)

It is one of the smallest of sea turtles and adults weigh about 45 kilograms and reach about 2 feet in length, their oval carapace is measures about the same in length and width, and is usually olive-green in color. The species, shown in Fig. 5, features a triangular-shaped head with two pairs of prefrontal scales and a somewhat hooked beak with large surfaces for crushing prey. This supports its shallow benthic diet [31] consisting primarily of crabs [32]. The Kemp's Ridley is the most endangered of all sea-turtle species [33].

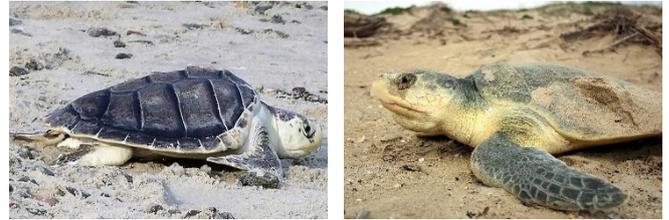

Fig. 5. Kemp's Ridley sea turtles.

## F. Olive ridley (Lepidochelys olivacea)

This is the smallest and most abundant [34] of all sea turtle species across the globe, an adult and a hatchling shown in Fig. 6. They are primarily found in the warm waters and despite their large numbers, their population decline has got them a vulnerable status in the IUCN Red List. They are carnivores [35] with their diet primarily composed of fish and crustaceans. Their curved carapace length is usually 58-77 cm and weigh about 35 kilogram on average, very seldom crossing the 45 kilogram mark [36]. When they aggregate in arribadas for mating [37], they were and are often susceptible to large-scale poaching for meat, eggs and skin [38].

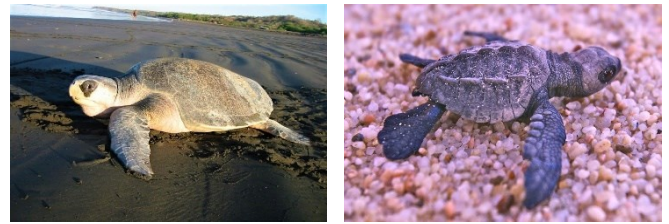

Fig. 6. Olive Ridley sea turtles.

## G. Flatback (Natator depressus)

The flatback turtles are endemic to the shallow turbid waters in the Australian continental shelf [39] and get their name from having a flattened carapace curvature than other sea turtles [40], as show in Fig. 7. The carapace is usually olive-green to grey and the plastron is cream colored. The IUCN Red List marks this species as Data Deficient, indicating insufficient information to indicate its current conservation status. They have the smallest geographic range of all sea turtle species and weigh about 70 kilograms.

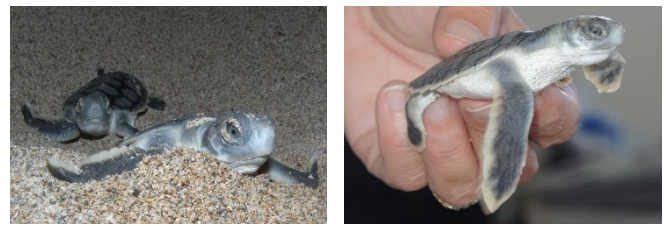

Fig. 6. Flatback sea turtles.

## III. CURRENT CONSERVATION STATUS

The status of the sea turtle population, despite current conservation efforts can be observed to be mostly Endangered in Table 1. This serves as a barometer to identify focus areas and species. For this analysis, the reports from the International Union for Conservation of Nature (IUCN) [41] in its Red List of Threatened Species, and from the Environmental Conservation Online System (ECOS) in accordance with the Endangered Species Act (ESA) of 1973 in the United States (elaborated in [42]) are considered.

Table 1. Lastest conservation status for all the Sea Turtles.

| Species | IUCN Status | ESA Status |
| --- | --- | --- |
| Loggerhead | **VU** [43] | **E+T** [44-46] |
| Green Turtle | **EN** [47] | **E +T** [48,49] |
| Leatherback | **VU** [50] | **E** [51,52] |
| Hawksbill | **CR** [53] | **E** [54,55] |
| Kemp's Ridley | **CR** [56] | **E** [57,58] |
| Olive Ridley | **VU** [59] | **E+T** [60,61] |
| Flatback | **DD** [62] | - |

For reference, the IUCN divides species into 9 categories: Not Evaluated, Data Deficient, Least Concern, Near Threatened, Vulnerable, Endangered, Critically Endangered, Extinct in the Wild and Extinct. The ESA divides species into 5 categories: Endangered (E), Threatened (T), Candidate (C), Endangered or Threatened due to similarity of Appearance (E(S/A) or T(S/A)), and Experimental Essential or Non-Essential (XE or XN). More information about these categories is outside the scope of this survey but can be gleaned from [41,42].

In Table 1, some species have multiple ESA protection statuses, indicating the variation in status across distinct population segments (DPS) of the species around the globe. Leatherbacks, Hawksbills and Kemp's Ridleys are endangered across their range. Flatbacks have not yet been surveyed extensively under the ESA standards and the remaining species of sea turtles are all endangered in some DPSs and threatened in the rest. The details about individual DPSs, conservation efforts and results are discussed in the status reports and 5 yearly reviews under ESA in Table 1.

## IV. FACTORS IMPACTING SEA TURTLES

Sea Turtles face a huge number of threats, which include but are not limited to illegal poaching of sea turtles for eggs, meat and more, shell trade, commercial fishing, especially longline trawl and gillnet fisheries, ingestion and entanglement in marine debris, artificial lighting, coastal armoring, beach erosion, and various beach activities like driving, putting up furniture, as well as predation by invasive species, marine pollution, oil spills and global warming. This section will examine these threats and their impact on sea turtle population in detail.

Poaching, though illegal, is still very much a threat to sea turtles, with high mortality of loggerheads having been reported in the coastal waters of Baja California Sur (BCS) [63] due to bycatch, human consumption and stranding. In another analysis [26] of 1945 carcasses in Bahia Magdalena, on the Pacific coast of BCS, human consumption constituted 95-100% of the deaths in towns, whereas on beaches, 76-100% were of unknown causes, though circumstantial evidence suggested them being incidental bycatch. Incidental capture of sea turtles continue across the world, with a lot of reports and analysis from around Brazil [64,65], Canada [66], Peru [67], Chile (south-eastern Pacific) [68], tropical northeast Atlantic [69] and India [70]. The Convention on International Trade in Endangered Species of Wild Fauna and Flora (CITES) banned all international trade in marine turtles, and has massively helped global conservation efforts.

Plastic is known to cause massive menace to sea turtles and the entire marine ecosystem; however, researchers are yet making advances into understanding the various pathways of this damage. Previously it was reported that sea turtles consumed plastic bags due to their visual similarity to jellyfish [71-74], now it has also been suggested that in addition to the aforesaid interactions, odors from marine plastics also elicit foraging behavior in [75], which means that turtles would come far away to these 'olfactory traps' and their normal foraging strategies could be detrimental or even lethal.

The plastic threat however is more widespread and also exists on sea beaches and includes fishing nylon lines, rope fragments and other anthropogenic products, rope fragments etc. In 2011, postmortem analysis of 23 specimens of 5 species in Brazil indicated 39% of them had consumed soft and hard plastic, polyethylene terephthalate (PET) bottle caps, human hair, metal fragments, latex condoms and tampons [76]. Consumption and entanglement in plastic has been reported to cause reduced feeding, movement, reproduction, injuries, ulcers and lacerations and ultimately death [77,78].

Artificial light is detrimental for sea turtles and hatchlings in a variety of ways. A lot of literature including [79,80] indicate that anthropogenic light pollution can cause sea turtle hatchlings to lose their sense of orientation on their journey towards the sea. [81] provided the first experimental evidence proving turtles spend more time near-shore even after reaching water with beach lighting influences, with 88% of the trajectories oriented towards the light source. Their work took into account the current speeds and the observations held true even when current direction was offshore. This behavior subjects the turtles to greater threats of predation. [82] studied green turtles, leatherbacks and loggerheads and found significant negative correlation between light pollution and nest density for all the species under study, they also highlighted that this anthropogenic environmental pressure was most pronounced amongst green turtles, followed by leatherbacks and loggerheads in that order.

Oil spills are not the most significant threat to sea turtles and are comparatively rare occurrences but have a catastrophic impact on the population when they happen as indicated in [83-85], not only affecting the sea turtles directly, but also leading to massive habitat degradation. The Deepwater Horizon Oil Spill which spewed unprecedented amounts of oil and dispersants into hard to study pelagic environments [86] was associated with the prevalence of 1.5 times more total and late-stage deformities in Kemp's Ridley embryos, with the most common being craniofacial and carapace deformities. In two areas under study in Texas, the number of documented nests also decreased after the spill, after an exponential rise starting from 1995 [87].

Global warming plays a very important role in the lost term sustainability of sea turtle populations as it has been extensively experimented [88] and well-documented [89,90] that offspring are temperature-biased, with predominantly female offspring in warm climates and male offspring in cool climates. Warm temperatures are also associated with higher embryo mortality [91]. With sea turtle ratios continuing to slide towards the female sex with global warming, [92] explores better managements techniques with strategic cooling of the eggs, helping long term species preservation.

Beach armoring [93-96] and nourishment have also been shown to be detrimental to nesting sea turtles and hatchlings. Beach nourishment can affect nesting success [97] by causing beach compaction, changing the nest-chamber geometry and concealment etc, and also the escarpments make it harder for females to reach nesting areas. Other beach activities [98] like beach driving and furniture can cause sand compaction, tire ruts can cause ridges hard to surmount for hatchlings while trying to reach the sea. Vehicles, furniture and mechanical cleaning of beaches can also crush existing eggs. Obstructions laid on the beach can make it harder for hatchlings to reach the sea and also increase predation rates. Additionally, litter and garbage draw non-native predators like raccoons, mongoose and dogs that eat eggs and hatchlings.

This section has covered the various factors that impact sea turtle populations across the globe, further discussion into the various anthropogenic pressures can be found in [99]. The following section will analyze how ML and Computer Vision play an important role in bolstering sea turtle conservation efforts through real time detection, identification and more.

V. MACHINE INTELLIGENCE ADVANCES IN DETECTION AND IDENTIFICATION OF SEA TURTLES

Convolutional Neural Networks (CNNs) have gained immense popularity in computer vision research, including but not limited to recognition and identification tasks. Liu et al. [100] demonstrated a turtle recognition and identification system based on CNNs and Transfer Learning. After comparing various approaches in that domain, Indecption-V3 with Transfer Learning was reported to have the best performance with a validation accuracy of 96.83%.

Drone imagery is also useful in conducting safe, cost-effective and robust abundance assessments as demonstrated in this manual pilot study [101]. CNNs were also reported to have been efficient in identifying turtles in drone imagery from Unoccupied Aircraft Systems (UAS) in [102] and exceeded human accuracy by a difference of 8.6% on the database introduced in [101].

Faster R-CNNs were used in detecting sea turtles in [103], using images from either static cameras, moving robots or airborne drones and processing them in the cloud, the processing had a speed of 3 frames per second. [104] presented two algorithms for classification of two sea turtle species of the Mediterranean region, namely loggerhead and green turtles. The first method involved a Bag-of-Words model with a Linear SVM achieving 69% accuracy, whereas the CNN classifier achieved 70.14% accuracy, after 20 epochs using a 5x5 filter. Deng et al. [105] proposed an energy efficient system to detect sea turtles in image and video feeds, targeted at fishing boats, with an aim to reduce bycatch.

Interestingly, Computer Vision is not limited to just camera images and [106] presented an autonomous system for sea turtle detection using a multibeam sonar and CNNs, which is suitable for Remotely Operated Vehicles (ROV) or Autonomous Underwater vehicles (AUV). The paper reported an average precision of 75%. Horimoto et al. [107] presented a way to track sea turtles under water and chase them with an AUV, using a CNN and multibeam sonar approach as well.

Photo-identification (PID) of the individuals within a species is an active research area. In 1973, Ekambaram and Kasturi [108] developed a PID system that leveraged SIFT feature detection and Cumulative Match Characteristics (CMC) curve, capturing the identification probability of various ranks. The algorithm was tested with a gallery of 779 images and a probe of 114 images. They reported a matching image appearing in the top 10% of the gallery for 70% of the probe images and coverage increasing to 80% if top 20% of the gallery is considered.

Reisser et al. [109] were the first to use facial scute shapes and arrangement patterns to identify sea-turtles and report the viability of the technique for a period of a minimum 1046 days. Schofield et al. [110] demonstrated the effectiveness of using natural markings occurring on the post-ocular regions of sea turtle faces and achieved 68-100% mean accuracy in all trials on a Loggerhead sea turtle population exceeding 400 individuals. Further they noted its suitability for male turtles who rarely come ashore, rendering conventional slipper tagging unsuitable. Their research facilitated better assessment of adult sex ratios through broader identification.

Jean et al. [111] proposed another photo-identification method for green and hawksbill turtles based on the facial profile and shape of the scutes. The primary shortcoming was maintaining the region of identified scales, even minor variations to which would massively impact the identification rate. In 2014, Carter et al. [112] developed a PID system using artificial neural networks called MYDAS, for green turtles (Chelonia mydas) and reported an accuracy greater than 95%, which was deployed on Lady Elliot island in the southern part of the great barrier reef aiding conservation efforts. Carpenter et al. [113] validated facial scutes based photoidentification techniques by reporting the stability of the facial scale patterns for periods upto 11 years, despite minor changes in pigmentation.

In recent times, a lot more work has been done in photoidentification using facial scutes like [114] for green turtles, [115] for juvenile hawksbills, and assessments using the I3S software [116] (Interactive Individual Identification System software) for free swimming green turtles and hawksbills in [117] and nesting hawksbills in [118]. Photo identification of a dead green turtle was also reported in [119].

Rao et al. [120] proposed a PID system using map graphs of the scales on sea turtles' faces. In 2021, Dunbar et al. [121] demonstrated another PID system called HotSpotter, for a database of 2136 images of Hawksbill turtles and a set of 158 query images, it had a top1 accuracy of 80% and top-6 accuracy of 91% despite varying underwater conditions, angles and image quality. HotSpotter was integrated into the Wildbook [122] based Internet of Turtles [123,124].

Gatto et al. [125] proposed a novel system of identifying sea turtles using the unique patterns on their front flippers with many added advantages, like the flippers having more scutes, not requiring to disturb the nesting females with bright light flashes and the ability to identify hatchlings both over a short and long term, which was not possible with facial patterns or conventional tagging . It was tested on green turtles and olive ridley turtles and reported an accuracy of 92.9% and 81.8% respectively. PID advances like these have a massive impact in the long-term capture-mark-recapture (CMR) studies which are the 'gold standard' for conservation [126], providing robust population assessments and trend analysis.

Okuyama et al. [127] applied template matching to analyze horizontal head movements in an underwater environment, they reported it being advantageous to sensors as no analysis was needed to acquire information. The method fails when the turtles head moves out of the camera's field of view which depends on the camera position and field of view. Despite its limitations, this is the first effective demonstration of bio-logging with computer vision for sea turtles and can give a new perspective to understanding turtle movements.

## VI. Ongoing Research

The author of this work is currently developing algorithms for robust and scalable sea turtle face detection in diverse environments and is ranked third on the public leaderboard for the international challenge on *Zindi* titled 'Local Ocean Conservation – Sea Turtle Face Detection' [128]. The algorithm will soon be published for dissemination across the Wildlife Conservation and Computer Vision communities.

## VII. Future Research Avenues

This survey has threats and conservation status of the sea turtles and discussed the advances of ML and CV in various aspects of this conservation research. One avenue with significant scope would involve running these autonomous systems on edge devices leveraging TinyML, which has already proven to be useful [129] in other Computer Vision research areas like face-mask detection[130], underwater marine life and plastic waste detection [131], American Sign Language recognition [132], drone based detection and tracking [133], and diabetic retinopathy prognosis [134] etc.

Drone imagery can massively improve the analysis and response time, and has proven immensely useful in many fields [135], an interesting research avenue can be building and maintaining high quality image and video databases for sea turtles, on the beaches and underwater, helping create robust CV applications for detecting and identifying turtles, while also analyzing DPS population densities [136]. Research could also be done in the development of better algorithms to detect and identify turtles using their natural features and improving explainability of the systems.

## VIII. Conclusion

This survey comprehensively presents the current status of sea turtle conservation across species and geographies, their threats and remedies, and discusses some recent Computer Vision advances aiding conservation efforts. It also presents some future research avenues, helping conservation experts and Computer Vision researchers on a combined front to help rebuild sustainable sea turtle populations across the globe.

## Photo Credits and Acknowledgment

A number of images were used as part of the analysis and critical commentary of the characteristics, nesting habits, threats and conservation of sea turtles, which constitutes fair use. The author gratefully acknowledges the image creators, in Table 2 below, listing the creators, titles and licenses per image; (a, b) represent the (left, right) images per figure.

Table 2. Image Source Atributions

| Fig | Author | Title | License |
|---|---|---|---|
| 1a | U. Kanda | Loggerhead turtle | CC BY 2.0 |
| 1b | B. Gratwicke | Loggerhead Sea Turtle (Caretta caretta) | CC BY 2.0 |
| 2a | P. Asman, J. Lenoble | Chelonia mydas Green Sea Turtle | CC BY 2.0 |
| 2b | Maritime Aquarium at Norwalk | Aquar.green sea turtle.7 | CC BY-ND 2.0 |
| 3a | R. David, USFWS | Dermochelys coriacea | CC0 |
| 3b | USFWS Southeast Region | Leatherback sea turtle/ Tinglar, USVI | CC BY 2.0 |
| 4a | prilfish | Hawksbill Sea Turtle | CC BY 2.0 |
| 4b | C. S. Rogers | Hawksbill Sea Turtle/ Carey de Concha | CC0 |
| 5a | NER Sea Turtle Stranding Network | Kemp Ridley's sea turtle | Public Domain |
| 5b | via pxhere.com | Kemp's Ridley | CC0 |
| 6a | B. Flickinger | Olive Ridley sea turtle after laying eggs | CC BY 2.0 |
| 6b | S. Jurvetson | Releasing Baby Turtles to the Sea | CC BY 2.0 |
| 7a | S. Long, AIMS | Turtle Hatchlings | CC BY 3.0 AU |
| 7b | AIMS | A flatback turtle hatchling | CC BY 3.0 AU |